\documentclass[11pt]{article}

\usepackage[margin=1in]{geometry}
\usepackage{amsmath,amssymb,amsthm}
\usepackage{graphicx}
\usepackage{booktabs}
\usepackage{url}
\usepackage{hyperref}
\usepackage[numbers]{natbib}   
\usepackage{doi}              
\usepackage{color}            
\usepackage{bm}
\usepackage{enumitem}
\usepackage{listings}
\usepackage{xcolor} 
\usepackage{graphicx}
\usepackage{multicol}
\usepackage{setspace}
\usepackage{stfloats}

\usepackage{natbib}

\hypersetup{
    colorlinks=true,   
    linkcolor=black,   
    citecolor=blue,    
    filecolor=magenta, 
    urlcolor=blue      
}

\title{ProRCA: A Causal Python Package for Actionable Root Cause Analysis in Real-world Business Scenarios 

}

\author{\small
    Ahmed Dawoud\thanks{Head of Decision Science, ProfitOps; \texttt{adawoud@profitops.ai}} \quad
    \and 
    \small Shravan Talupula\thanks{Founder and CEO, ProfitOps; \texttt{stalupula@profitops.ai}}
}

\date{}

\lstdefinestyle{mystyle}{
    backgroundcolor=\color{gray!5},  
    commentstyle=\color{gray},        
    keywordstyle=\color{blue},        
    stringstyle=\color{red},          
    basicstyle=\ttfamily\scriptsize, 
    breakatwhitespace=false,         
    breaklines=true,                  
    captionpos=b,                     
    keepspaces=true,
    numbers=none,                      
    numberstyle=\tiny\color{gray},     
    showspaces=false,
    showstringspaces=false,
    showtabs=false,
    frame=single,                      
    tabsize=3,                         
    rulecolor=\color{gray},           
}

\begin{document}

\maketitle
\begin{abstract}
Root Cause Analysis (RCA) is becoming ever more critical as modern systems grow in complexity, volume of data, and interdependencies. While traditional RCA methods frequently rely on correlation-based or rule-based techniques, these approaches can prove inadequate in highly dynamic, multi-layered environments. In this paper, we present a pathway-tracing package built on the \emph{DoWhy} causal inference library. Our method integrates \emph{conditional anomaly scoring}, \emph{noise-based attribution}, and \emph{depth-first path exploration} to reveal multi-hop causal chains. By systematically tracing entire causal pathways from an observed anomaly back to the initial triggers, our approach provides a comprehensive, end-to-end RCA solution. Experimental evaluations with synthetic anomaly injections demonstrate the package’s ability to accurately isolate triggers and rank root causes by their overall significance.
\end{abstract}

\begin{multicols}{2}
\section{\large Introduction}
\label{sec:intro}
Modern operational landscapes, spanning domains such as retail, healthcare, finance, and software systems, are increasingly characterized by complex interdependencies and massive data streams. In such settings, anomalies rarely arise from a single isolated factor; rather, they emerge as the cumulative effect of multi-hop causal chains. Existing RCA methods typically focus on detecting outliers or isolating single nodes based on correlation or localized attribution. However, these approaches do not provide \emph{a complete} explanation of why a failure occurred. In other words they do not systematically trace \emph{all} possible causal pathways from an observed effect back to its initial triggers.

The primary motivation for our work is to address this limitation by developing a package that systematically reconstructs the full causal pathway from an observed anomaly back to its root cause. By leveraging the strengths of the \emph{DoWhy} causal inference library, our method extends existing techniques to not only identify individual anomalous nodes but also trace entire multi-hop causal chains. This end-to-end approach enables practitioners to intervene precisely at the earliest disruption points, thereby reducing the risk of recurring failures and improving overall system reliability.

\section{\large Literature Review}

Root Cause Analysis (RCA) is crucial for maintaining the reliability and performance of complex systems, spanning diverse domains~\cite{liu2021microhecl, wang2023interdependent, wang2023incremental, strobl2022identifying, okati2024root, budhathoki2021distribution, blobaum2022dowhy, chen2023automatic, jeyakumar2019explainit, strobl2024identifying, han2023root}.  However, existing RCA methodologies exhibit significant variations in their approach to causality, ranging from purely correlational techniques to those grounded in formal causal inference. This review categorizes and critically assesses these approaches, highlighting their strengths, limitations, and, crucially, whether they address the problem of tracing \textbf{all} root cause pathways. We identify three primary methodological groupings.

\subsection{\normalsize Correlation-Based Approaches}

Methods in this category primarily rely on statistical correlations between metrics, often combined with graph structures representing system topology or communication patterns, but they do \textbf{not} explicitly model causal mechanisms or employ formal causal inference techniques (e.g., do-calculus, interventions). For instance, \textbf{MicroRCA}~\cite{wu2020microrca} and \textbf{MicroHECL}~\cite{liu2021microhecl} fall squarely within this category by leveraging service call graphs and correlating performance symptoms with resource utilization. While these approaches are practical and scalable, they are susceptible to spurious correlations and may miss root causes that do not manifest in obvious resource spikes. Similarly, \textbf{ExplainIt!}~\cite{jeyakumar2019explainit} uses attributed graphs and correlations, making it weakly causal at best, as it explores a causality graph but the core methodology remains correlational. In another example, \textbf{Chen et al. (2023)} \cite{chen2023automatic} explores the use of Large Language Models (LLMs) for root cause analysis; however, this approach is weakly causal because it primarily depends on correlations and prompt engineering, thus being vulnerable to spurious relationships and potentially missing subtle causal pathways. Likewise, \textbf{Wang et al. (2023) - REASON}~\cite{wang2023interdependent} introduces interdependent causal networks and uses a hierarchical GNN. Despite the term "causal networks," the core methodology is grounded in learning correlations and propagating information through the network, and although the GNN learns representations, it does not explicitly model interventions or counterfactuals in a formally causal manner.

These methods are often efficient and applicable to large-scale systems. However, their reliance on correlation makes them vulnerable to confounding and limits their ability to provide robust, causally sound explanations. They do \textbf{not} aim to trace all root cause pathways.

\subsection{\normalsize Hybrid Approaches: Causal Discovery with Limitations}

Methods in this group incorporate elements of causal discovery but have significant limitations that prevent them from being considered fully causal. For example, \textbf{PORCA}~\cite{gong2024porca} uses "magnified structural causal models" and score-based causal discovery, which is a step toward genuine causal inference. Nevertheless, its effectiveness depends heavily on the novel score function and scheduling process, and the extent to which these truly capture causal relationships requires further scrutiny. Meanwhile, \textbf{Wang et al. (2023) - Incremental} \cite{wang2023incremental} employs causal discovery with trigger points; although this is indeed a causal approach, its scalability is constrained by the computational cost of causal discovery, and it does not aim for complete pathway tracing. Additionally, \textbf{Okati et al. (2024)} \cite{okati2024root} explores a traversal-based approach in combination with a causal graph model and Shapley values, making it weakly causal.

These methods represent a move toward causal reasoning, but they often rely on simplifying assumptions (e.g., about the form of causal relationships) or have limitations in scalability or completeness. They typically aim to \textbf{identify} or \textbf{localize} root causes, but not to trace all pathways.

\subsection{\normalsize Genuinely Causal Approaches}

The final group includes methods that are firmly grounded in formal causal inference frameworks, such as Structural Causal Models (SCMs) and do-calculus. \textbf{Budhathoki et al. (2021)}~\cite{budhathoki2021distribution} explicitly use SCMs and Shapley values to quantify the "Intrinsic Causal Contribution" (ICC) of changes in causal mechanisms, offering a theoretically sound approach but requiring the causal graph (DAG) to be known, which is a strong assumption in practice. \textbf{Bl\"obaum et al. (2022)}~\cite{blobaum2022dowhy} (DoWhy-GCM) presents a software package for causal inference, supporting various causal queries based on SCMs; however, it does not prescribe a specific method for root cause \textbf{\emph{pathway}} tracing. Additionally, \textbf{Strobl and Lasko (2022, 2024)}~\cite{strobl2022identifying, strobl2024identifying} introduce the heteroscedastic noise model (HNM) and Generalized Root Causal Inference (GRCI), which is genuinely causal but focuses on identifying patient-specific root causes in a medical context and does not address the \textbf{\emph{all pathways}} problem. Finally, \textbf{Han et al. (2023)} \cite{han2023root} proposes RootCLAM, a framework that uses a causal graph autoencoder.

These methods offer the strongest theoretical guarantees regarding causality. Crucially, none of these methods explicitly address the problem of tracing all possible root cause pathways.


\subsection{\normalsize Genuine Causal Pathway Tracking Approaches}

The PyRCA package is the only package we found that tracks complete causal pathways. It incorporates a spectrum of methods, some firmly grounded in causal inference theory, while others rely more on statistical correlations. For example, techniques like epsilon‑Diagnosis and random walk are primarily driven by changes in metric values and graph-based heuristics, respectively. These approaches flag anomalies based on statistically significant deviations or network connectivity but do not explicitly model interventions or counterfactuals. As a result, their causality claims are limited; they tend to capture correlations rather than true causation.

In contrast, methods such as the Bayesian inference approach and the hypothesis‑testing (CIRCA) algorithm explicitly leverage the known causal structure to assess how interventions propagate through the system. These methods build and utilize a structural causal model that mirrors the true data-generating process, evaluating whether observed changes in metrics can be directly attributed to a disruption at a specific node. When the underlying assumptions hold—such as the accuracy of the DAG, the absence of hidden confounders, and appropriate model specifications—these methods provide a more rigorous, truly causal analysis of root causes.

However, according to the package’s GitHub repository, all these causal methods from PyRCA have proven to be inaccurate—except for hypothesis‑testing (HT), which performed perfectly on the simulated dataset. Yet, HT is limited to continuous variables and uses only linear regression. In contrast, DoWhy employs nine different models for each node and selects the best one, allowing for robust fitting of causal mechanisms and accommodating varying levels of complexity in modeling relationships across different data types. Theoretically, DoWhy offers a more robust approach; therefore, extending it should enable us to achieve better results.


\section{\large Overall Gap and Positioning of the Proposed Work}

The Literature review revealed a significant gap in the existing literature. While many methods exist for RCA, most rely on correlations or make simplifying assumptions that limit their causal validity. Even the genuinely causal approaches do not focus on the exhaustive tracing of all possible root cause pathways, which is the core contribution of the proposed work. By addressing this gap, the proposed technique offers a novel and potentially more robust approach to RCA, particularly in complex systems where multiple interacting factors may contribute to a failure. 

 We argue that complete pathway tracing is essential for robust RCA.  Unlike methods that pinpoint a single failure point, our approach seeks to reveal the full \textbf{sequence} of causal events, providing a comprehensive understanding of the failure's etiology. This holistic view uncovers hidden interactions and dependencies, including feedback loops, that might be missed by localized analyses.  
 
 Furthermore, a complete pathway reveals multiple potential intervention points, enabling earlier and potentially less disruptive mitigation strategies.  By considering all pathways, our analysis is less susceptible to noise and spurious correlations, leading to a more reliable and, crucially, more \textbf{explainable} diagnosis. In essence, while identifying a root cause offers a valuable starting point, tracing \textbf{all pathways} provides the complete, robust, and actionable understanding necessary for effective failure prevention in complex systems, clearly distinguishing our approach.


\section{\large Methodology}

Our proposed package for root cause pathway analysis leverages core functionalities from the DoWhy Package—such as Structural Causal Model (SCM) construction, noise attribution, and conditional anomaly scoring—and extends them with novel contributions designed to extract, trace, and rank multi-hop causal pathways. In the section, we first briefly describe the inherited components and then focus on detailing our original contributions.

\subsection{\normalsize Inherited Components from DoWhy}

The baseline components used in our package are as follows:

\begin{itemize}
    \setlength{\itemsep}{\baselineskip} 
    \setlength{\leftmargin}{2em} 

    \item \textbf{SCM Construction:} We build a directed acyclic graph (DAG) \(G = (V, E)\), where \(V\) is the set of variables (nodes) and \(E\) is the set of causal edges. Each node \(v \in V\) is assigned a causal mechanism \(M(v)\) that maps parent inputs to its output. These steps—including automatic node inference, mechanism assignment, and model fitting—are implemented using DoWhy’s standard routines.

    \item \textbf{Noise Attribution:} For each node \(v\), a noise contribution vector \(C(v) \in \mathbb{R}^{n_v}\) is computed using DoWhy’s \texttt{attribute\_anomalies} function with a \texttt{MedianCDFQuantileScorer}. This vector quantifies the deviation of observed values from their expected distribution.

    \item \textbf{Conditional Anomaly Scoring:} 
    For nodes with parents, conditional anomaly scores \(S(v)\) are derived by comparing the observed value \(x_v\) against the prediction of the mechanism \(M(v)\) given the parent values \(\bm{x}_{P(v)}\). These scores are computed using DoWhy’s 
    \texttt{anomaly.\allowbreak conditional\allowbreak\_anomaly\allowbreak\_scores} function.

\end{itemize}

\subsection{\normalsize Our Contributions}

Our work extends the DoWhy package with the following novel components:

\subsubsection{\small Combined Node Score}

We introduce a \emph{combined node score} that fuses the structural anomaly score and the noise contribution into a single metric for each node \(v\). Specifically, let:
\begin{enumerate}[label=(\alph*)]
    \item The average structural anomaly score over the anomaly subset \(S_A(v)\) be
    \begin{equation}
    \bar{S}(v) = \frac{1}{|S_A(v)|} \sum_{s \in S_A(v)} s. \tag{1}
    \end{equation}
    
    \item The maximum absolute noise contribution be
    \begin{equation}
    \tilde{C}(v) = \max_{i=1,\dots,n_v} \Big\{ \big| C(v)_i \big| \Big\}. \tag{2}
    \end{equation}
\end{enumerate}
Our combined score is then defined as a weighted sum:
\begin{equation} \label{eq:combined_score}
\text{CombinedScore}(v) = \alpha \, \bar{S}(v) + (1-\alpha) \, \tilde{C}(v), \tag{3}
\end{equation}
where \(\alpha\) is a weight parameter (set to \(\alpha = 0.7\) in our experiments). Equation \eqref{eq:combined_score} represents our primary novel contribution, effectively integrating both model-driven and empirical anomaly information.

\subsubsection{\small Root-Cause Path Discovery}

We propose a depth-first search (DFS) algorithm to trace multi-hop causal pathways from a target node (e.g., \texttt{PROFIT\_MARGIN}) back to potential root causes. Let the target node be denoted by \(v_0\) and consider a path:
\[
P = \{v_0, v_1, \dots, v_k\},
\]

where each \(v_{i+1}\) is a parent of \(v_i\). Our algorithm extends a path based on the following threshold criteria:

\begin{enumerate}[label=(\roman*)]
    \item For each intermediate node \(v_i\) (\(i = 1, \dots, k-1\)), require:
    \begin{equation} \label{eq:intermediate_threshold}
    \text{CombinedScore}(v_i) \geq \beta \theta, \tag{4}
    \end{equation}
    where \(\theta\) is a predefined threshold (e.g., \(\theta = 0.8\)) and \(\beta\) is set to \(0.7\).
    
    \item For the terminal (root) node \(v_k\), require:
    \begin{equation} \label{eq:terminal_threshold}
    \text{CombinedScore}(v_k) \geq \theta. \tag{5}
    \end{equation}
\end{enumerate}

Thus, the set of accepted paths is defined as:

\begin{equation} \label{eq:accepted_paths}
\resizebox{0.9\hsize}{!}{$
\mathcal{P} = \left\{ P = \{v_0, v_1, \dots, v_k\} \,\Bigg|\, 
\begin{aligned}
&\forall\, i=1,\dots,k-1:\, \text{CombinedScore}(v_i) \geq \beta \theta, \\[1mm]
&\text{and} \quad \text{CombinedScore}(v_k) \geq \theta
\end{aligned}
\right\}. \tag{6}
$}
\end{equation}

This DFS-based path discovery is our original extension, enabling the exploration of multi-hop causal chains based on a composite importance measure.

\subsubsection{\small Path Significance Evaluation}

To rank the candidate causal pathways, we introduce a novel metric that evaluates each path based on a combination of the root node's noise anomaly and the causal consistency along the path.

\paragraph{(a) Noise Component:}  
Let the average noise contribution at the terminal node \(v_k\) be
\begin{equation} \label{eq:noise_component}
\mu_{v_k} = \frac{1}{n_{v_k}} \sum_{i=1}^{n_{v_k}} C(v_k)_i. \tag{7}
\end{equation}

\paragraph{(b) Causal Consistency Component:}  
For each adjacent pair \((v_i, v_{i+1})\) along the path, compute the absolute Pearson correlation coefficient between their noise vectors:
\begin{equation} \label{eq:correlation}
\rho_{i,i+1} = \left| \operatorname{corr}\Bigl( C(v_i),\, C(v_{i+1}) \Bigr) \right|. \tag{8}
\end{equation}
Then, the overall consistency of the path is given by:
\begin{equation} \label{eq:consistency}
\operatorname{Consistency}(P) = \frac{1}{k} \sum_{i=0}^{k-1} \rho_{i,i+1}. \tag{9}
\end{equation}

\paragraph{(c) Overall Path Significance:}  
We compute the significance of the path as a weighted sum of the noise and consistency components:
\begin{equation} \label{eq:path_significance}
\operatorname{Significance}(P) = \gamma \, \mu_{v_k} + (1-\gamma) \, \operatorname{Consistency}(P), \tag{10}
\end{equation}
with \(\gamma\) set to \(0.7\). Equations \eqref{eq:noise_component}--\eqref{eq:path_significance} together constitute our novel metric for ranking multi-hop causal pathways.

\subsection{\normalsize Summary of the Methodology}

ProRCA's methodology integrates core functionalities of the DoWhy package with novel contributions for complete causal pathway analysis.  First, inheriting from DoWhy, we construct a Structural Causal Model (SCM), building a directed acyclic graph (DAG) \(G = (V, E)\) and assigning causal mechanisms \(M(v)\) to each node \(v\).  We then compute, for each node, a noise contribution vector \(C(v)\) and conditional anomaly scores \(S(v)\), also leveraging DoWhy's established procedures.

Building upon this foundation, ProRCA introduces a combined node score, fusing structural anomaly and noise contribution:  \(\text{CombinedScore}(v) = 0.7\,\bar{S}(v) + 0.3\,\tilde{C}(v)\). This single metric integrates both model-based and empirical anomaly information.  To trace root cause pathways, we perform a depth-first search (DFS) starting from the target node \(v_0\), identifying paths \(P = \{v_0, v_1, \dots, v_k\}\) that meet specific threshold criteria for both intermediate and terminal nodes (Equations \eqref{eq:intermediate_threshold} and \eqref{eq:terminal_threshold}). This yields a set of accepted paths, \(\mathcal{P}\) (Equation \eqref{eq:accepted_paths}).

Finally, we evaluate the significance of each discovered path \(P\) by combining the terminal node's average noise contribution (\(\mu_{v_k}\)) with a measure of causal consistency along the path (\(\operatorname{Consistency}(P)\)): \(\operatorname{Significance}(P) = 0.7\,\mu_{v_k} + 0.3\,\operatorname{Consistency}(P)\). This novel metric prioritizes paths with both significant root cause noise and strong causal coherence.  In essence, ProRCA extends DoWhy with a combined scoring system, a DFS-based pathway discovery algorithm, and a path significance evaluation, providing a comprehensive package for multi-hop root cause analysis.


\section{\large Complete ProRCA Workflow}
\label{sec:anomaly_injection}
This section details the process of anomaly injection, the implementation pipeline, and the evaluation metrics.  The complete code, including data generation scripts and analysis notebooks, is publicly available in the \emph{notebook} folder of our GitHub repository.\footnote{\url{https://github.com/profitopsai/ProRCA}}

\subsection{\normalsize Synthetic Data Generation} The synthetic dataset is produced by a Python-based simulation that approximates real-world retail transactions in an intricate, non-linear manner. 

\begin{lstlisting}[language=Python, caption={Example Data Generation Code 2023}, label={lst:data generation class}]
# Example usage
start_date = '2023-01-01'
end_date = '2023-12-30'
df = generate_fashion_data_with_brand(start_date, end_date)
\end{lstlisting}

At its core, the code assigns product attributes—such as category, brand, and hierarchical merchandise structure—to each generated record, reflecting a diverse range of items (e.g., \texttt{Apparel.Men.Shirts.Casual}, \texttt{Footwear.Women.Boots.Winter}). Transaction-level features exhibit pronounced non-linearity through lognormal draws for \texttt{PRICEEACH}, Poisson sampling for \texttt{QUANTITYORDERED}, and multiplicative modifiers for costs and discounts. Randomization of sales channels, geographic territories, and promotion codes further amplifies the complexity, while weekly seasonality (weekends vs. weekdays) influences base demand.

Key financial metrics, including $\texttt{SALES}$, $\texttt{DISCOUNT}$, $\texttt{NET\_SALES}$, $\texttt{FULFILLMENT\_COST}$, $\texttt{COST\_OF\_GOODS\_SOLD}$, $\texttt{MARKETING\_COST}$, and $\texttt{RETURN\_COST}$, are computed by non-linear functions that factor in brand/category parameters, loyalty tiers, and location-specific multipliers. 

This layered approach ensures that transactional outcomes—such as $\texttt{PROFIT}$ and $\texttt{PROFIT\_MARGIN}$—arise from realistic, interdependent processes. Specifically, the generation script recasts each row’s profit by combining net sales, shipping revenue, and cost components in a manner that imitates real-world operations, introducing the potential for complex interactions and non-trivial outliers.


\subsection{\normalsize Controlled Anomaly Injection} To rigorously assess ProRCA’s root cause detection, the code injects anomalies at specified dates and scopes—either delimited by $\texttt{SALES\_CHANNEL}$ or product hierarchy—using non-linear transformations. By recalculating every downstream metric within each affected transaction, the injection script ensures robust, interlinked disruptions that realistically resemble operational mishaps. Such non-linear perturbations impose additional challenges for detection, necessitating causal inference techniques capable of disentangling multi-hop dependencies. 

\begin{lstlisting}[language=Python, caption={Anomaly Injection Schedule}, label={lst:anomaly_schedule}]
anomaly_schedule = {
    '2023-01-10': ('Discount',0.6,'Apparel'),
    '2023-06-10': ('COGs', -0.8, 'Footwear'),
    '2023-09-10': ('Fulfill', -2,'Beauty'),
    '2023-12-10': ('Return',10,'Accessories')
}

df_anomalous = inject_anomalies_by_date(df, anomaly_schedule)
\end{lstlisting} 

Figure 1 shows four types of anomalies are introduced on different dates and product lines, each with a unique root cause, severity, and non-linear effect on key variables:

\textbf{ExcessiveDiscount (2023-01-10, Apparel):}  
Simulates a pricing error by boosting the discount to an unusually high proportion of $\texttt{SALES}$. This anomaly drastically lowers $\texttt{NET\_SALES}$ and propagates through downstream metrics, including $\texttt{RETURN\_COST}$ and $\texttt{PROFIT\_MARGIN}$.

\textbf{COGSOverstatement (2023-06-10, Footwear):}  
Models a supplier issue by artificially increasing $\texttt{UNIT\_COST}$, thereby raising $\texttt{COST\_OF\_GOODS\_SOLD}$. The resulting effect lowers $\texttt{PROFIT}$ and, in turn, reduces $\texttt{PROFIT\_MARGIN}$.

\textbf{FulfillmentSpike (2023-09-10, Beauty):}  
Represents a logistics problem that inflates $\texttt{FULFILLMENT\_COST}$ in a non-linear manner. This spike ripples through $\texttt{PROFIT}$ calculations, ultimately diminishing $\texttt{PROFIT\_MARGIN}$.

\textbf{ReturnSurge (2023-12-10, Accessories):}  
Emulates a quality control issue causing an abrupt jump in $\texttt{RETURN\_COST}$. This leads to marked reductions in $\texttt{PROFIT}$ and $\texttt{PROFIT\_MARGIN}$, with potential downstream effects on other revenue- and cost-related features. \\

\noindent\makebox[\columnwidth]{%
    \includegraphics[width=0.5\textwidth]{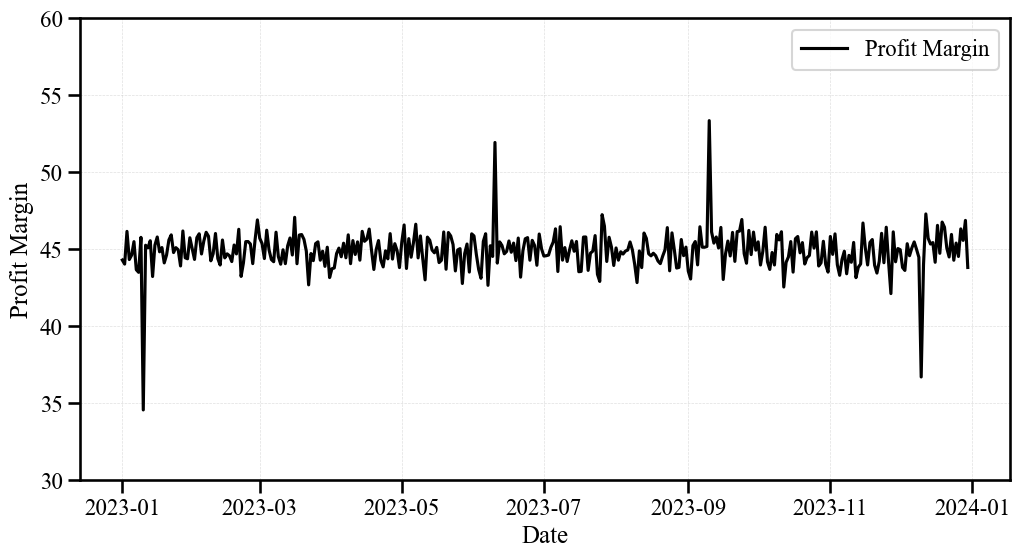}
}
\par
\noindent\textbf{\normalsize Figure 1:} Profit Margin Analysis Over Time.


\subsection{Anomaly Detection}
In the this step, the \texttt{AnomalyDetector} identifies dates where the \texttt{PROFIT\_MARGIN} metric deviates from expected patterns.  This module flags anomalous dates; it does not attempt to explain why the anomalies occurred.  The \texttt{AnomalyDetector} is implemented as a Python class, leveraging the `InterQuartileRangeAD` detector from the `adtk` library.  This detector is a standard, non-parametric method for identifying outliers in time series data.  It flags values that fall outside a dynamically calculated range based on the interquartile range (IQR) of the data.

The code for the \texttt{AnomalyDetector} is as follows:

\begin{lstlisting}[language=Python, caption={Anomaly Detection Code}, label={lst:anomaly_detector}]
detector = AnomalyDetector(df_agg, date_col="ORDERDATE", value_col="PROFIT_MARGIN")

anomalies = detector.detect()
anomaly_dates = detector.get_anomaly_dates()
detector.visualize(figsize=(12,6))
\end{lstlisting} 

The \texttt{AnomalyDetector} class provides three key methods:
\texttt{detect()}, \texttt{get\_anomaly\_dates()}, and \texttt{visualize()}. 
The \texttt{detect()} identifies anomalies in the specified \texttt{value\_col}. The \texttt{get\_anomaly\_dates()} method extracts 
the dates where anomalies were detected. The \texttt{visualize()} method plots 
the time series data, highlighting the detected anomalies in red, as shown in 
Figure 2.

\noindent\makebox[\columnwidth]{%
    \includegraphics[width=0.5\textwidth]{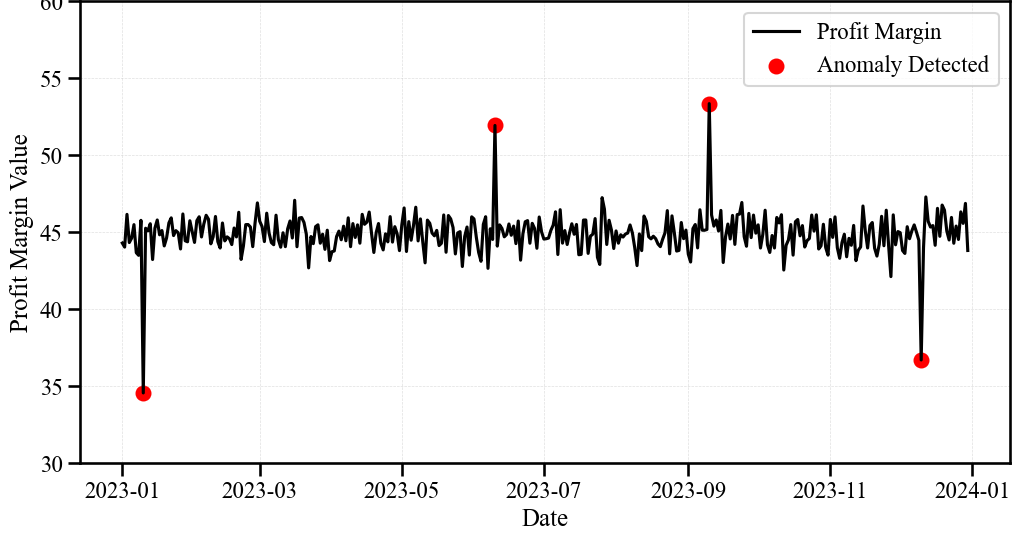}
}
\noindent\textbf{Figure 2:} Detected Anomaly Dates in Red.

Table 1 exemplifies the output of the anomaly detection, indicating that June 10, July 15, and October 30 are flagged as suspicious dates. \\

\noindent
\begin{minipage}{\columnwidth}  
\centering

\label{tab:detected_anomalies}
\renewcommand{\arraystretch}{1.1}
\small  

\begin{tabular}{p{0.55\columnwidth} p{0.3\columnwidth}} 
\toprule
\textbf{Date}       & \textbf{Suspicious?} \\
\midrule
2023-01-10          & Yes \\
2023-06-10          & Yes \\
2023-09-10          & Yes \\
2023-12-10          & Yes \\
\bottomrule
\end{tabular}
\vskip 8pt
\noindent{\normalsize \textbf{Table 1:} Initial Anomaly Detection} 
\end{minipage}


\subsection{Building the DAG and Fitting the SCM}

The \texttt{CausalRootCauseAnalyzer} operates on a Structural Causal Model (SCM) comprising a Directed Acyclic Graph (DAG) and corresponding causal mechanisms. In our controlled experiments, the DAG and functional forms are known, enabling direct construction of the SCM. Figure~3 shows the causal structure among the key variables in the data frame. These relationships reflect typical business logic in retail. \\

\noindent\makebox[\columnwidth]{%
    \includegraphics[width=0.5\textwidth]{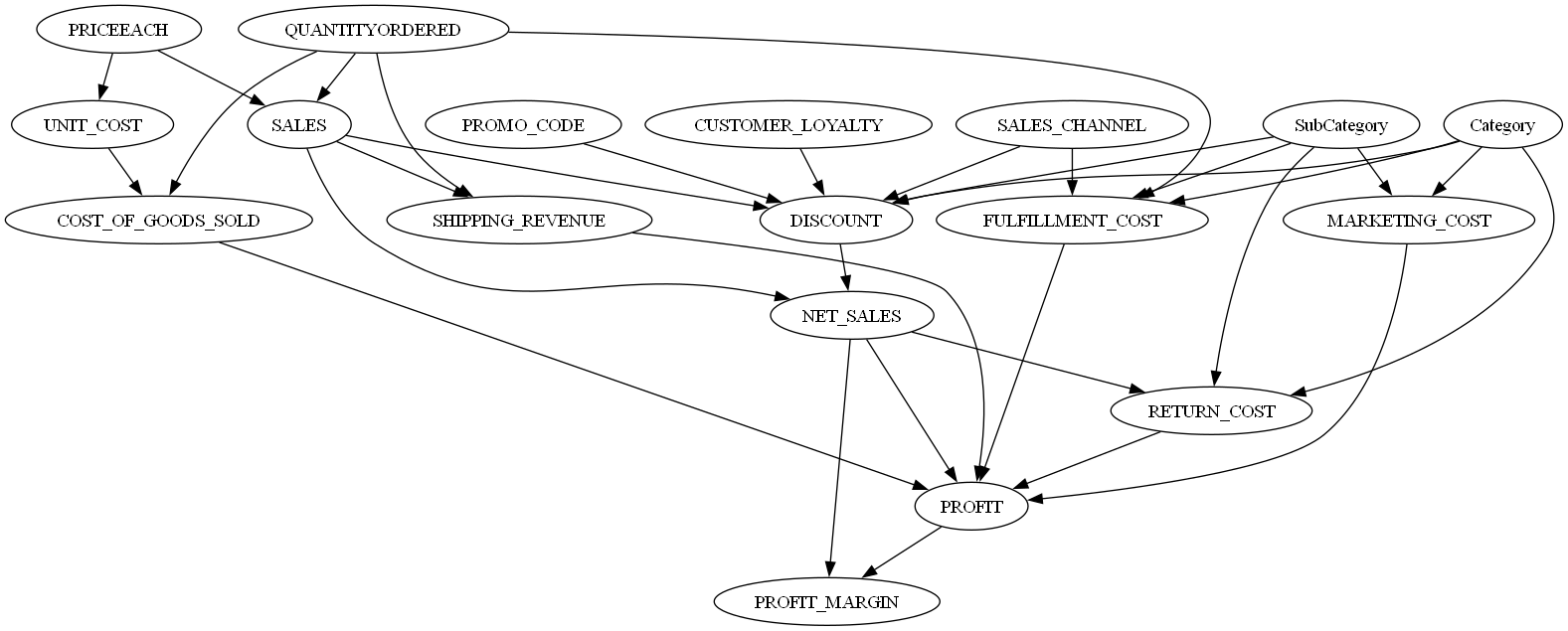}
}
\par
\noindent\textbf{\normalsize Figure 3:} Directed Acyclic Graph (DAG)

\vspace{5pt}

To construct the SCM, the \texttt{ScmBuilder} class takes a list of directed edges (e.g., \texttt{(PRICEEACH, SALES)} and assigns appropriate causal mechanisms to each node based on the data as illustrated in listing 2.

\begin{lstlisting}[language=Python, caption={SCM Building and Pathway Tracking}, label={lst:scm_building}, basicstyle=\footnotesize\ttfamily]
scm = ScmBuilder(edges=causal_edges)
scm.build(df_agg)
analyzer = CausalRootCauseAnalyzer(scm, min_score_threshold=0.3)
analysis_results = analyzer.analyze(df_agg, anomaly_dates)
\end{lstlisting}

\begin{raggedright}
Once the SCM is built, \texttt{\small CausalRootCauseAnalyzer} is initialized for anomaly analysis. Here, \texttt{min\_score\_threshold} governs the sensitivity of the root-cause pathways, determining which nodes qualify as anomalous. The entire pipeline, from DAG specification to root cause analysis, is thus encapsulated within the \texttt{CausalRootCauseAnalyzer} class. \end{raggedright} 

\subsection{Results: Root Cause Pathways}
The ProRCA package was evaluated on the detected anomaly dates from the step 5.2 (2023-01-10, 2023-06-10, 2023-9-10, and  2023-12-10). For each date, the system extracted a complete causal pathway from the outcome node (\texttt{PROFIT\_MARGIN}) back to a candidate root cause. The package computes two key metrics for each node: a \emph{Combined Score} (a weighted sum of the structural anomaly score and noise contribution) and the \emph{Noise Contribution} (the maximum absolute deviation from expected behavior). In addition, the overall \emph{Path Causal Significance} (PCS) is determined as a weighted combination of the terminal node's average noise and the causal consistency along the path.

The \texttt{min\_score\_threshold} parameter sets the minimum acceptable combined score for a node to be included in the DFS-based causal pathway. A higher threshold (e.g., 0.8) admits only nodes with very strong anomaly signals, while a lower threshold (e.g., 0.3) increases sensitivity by including nodes with lower scores. In our experiments, both settings yielded robust complete pathways; however, the high threshold ensured that only the most significant nodes were retained. The reported results below are based on a conservative threshold of 0.8

\subsubsection{\normalsize Accuracy of Pathway Identification}

Table 2 summarizes the causal pathways identified by ProRCA for the three anomaly dates. For each anomaly date (first column), the corresponding complete causal path (second column) is reported, along with the overall Path Causal Significance (PCS) in the third column. 

Notably, the pathway identified for 2023-01-10 is \texttt{PROFIT\_MARGIN} $\rightarrow$ \texttt{NET\_SALES} $\rightarrow$ \texttt{DISCOUNT}, which directly corresponds to the originally injected \emph{ExcessiveDiscount} anomaly. Similarly, the pathway for 2023-06-10, namely \texttt{PROFIT\_MARGIN} $\rightarrow$ \texttt{PROFIT} $\rightarrow$ \texttt{COST\_OF\_GOODS\_SOLD} $\rightarrow$ \texttt{UNIT\_COST}, aligns with the \emph{COGSOverstatement} anomaly that was introduced. The third pathway for 2023-9-10, \texttt{PROFIT\_MARGIN} $\rightarrow$ \texttt{PROFIT} $\rightarrow$ \texttt{FULFILLMENT\_COST}, reflects the impact of the injected \emph{FulfillmentSpike} anomaly. 

Finally, the third pathway for 2023-12-10, \texttt{PROFIT\_MARGIN} $\rightarrow$ \texttt{PROFIT} $\rightarrow$ \texttt{RETURN\_COST}, reflects the impact of the injected \emph{ReturnSurge} anomaly.  Thus, ProRCA successfully and accurately identified the correct multi-hop causal pathways, confirming its effectiveness in pinpointing the root causes corresponding to the pre-defined anomalies. \\

\noindent
\begin{minipage}{\columnwidth}
\centering
\label{tab:summary_paths}
\renewcommand{\arraystretch}{1.1}
\small
\begin{tabular}{p{0.20\columnwidth} p{0.70\columnwidth}}
\toprule
\scriptsize\textbf{Anomaly Date} & \scriptsize\textbf{Identified Causal Path} \\
\scriptsize 2023-01-10 & \scriptsize \texttt{PROFIT\_MARGIN} $\rightarrow$ \texttt{NET\_SALES} $\rightarrow$ \texttt{DISCOUNT} \\
\scriptsize 2023-06-10 & \scriptsize \texttt{PROFIT\_MARGIN} $\rightarrow$ \texttt{PROFIT} $\rightarrow$ \texttt{COST\_OF\_GOODS\_SOLD} $\rightarrow$ \texttt{UNIT\_COST} \\
\scriptsize 2023-9-10 & \scriptsize \texttt{PROFIT\_MARGIN} $\rightarrow$ \texttt{PROFIT} $\rightarrow$ \texttt{FULFILLMENT\_COST} \\
\scriptsize 2023-12-10 & \scriptsize \texttt{PROFIT\_MARGIN} $\rightarrow$ \texttt{PROFIT} $\rightarrow$ \texttt{RETURN\_COST} \\
\bottomrule
\end{tabular}
\vskip 10pt
\noindent{\small \textbf{Table 2:} Summary of Identified Causal Pathways}
\end{minipage}


\subsubsection{Detailed Node-Level Statistics}

\paragraph{The pathway extracted for 2023-01-10 consists of three nodes:} \texttt{PROFIT\_MARGIN}, \texttt{NET\_SALES}, and \texttt{DISCOUNT}. All nodes exhibit high Combined Scores (0.8526, 0.8267, and 0.8479, respectively), indicating that each node strongly contributed to the anomaly signal. Notably, the Noise Contribution increases along the causal chain—from 0.0379 at \texttt{PROFIT\_MARGIN} to 0.0998 at \texttt{NET\_SALES} and reaching 0.1241 at \texttt{DISCOUNT}. This trend suggests that while the outcome node already exhibits anomalous behavior, the anomaly becomes more pronounced upstream, with \texttt{DISCOUNT} displaying the highest noise level. Such a result is consistent with the pre-injected \emph{ExcessiveDiscount} anomaly, thereby confirming that ProRCA correctly identified the underlying causal pathway.\\

\noindent
\begin{minipage}{\columnwidth}
\centering
\label{tab:results_0610}
\renewcommand{\arraystretch}{1.1}
\small
\begin{tabular}{p{0.35\columnwidth} p{0.2\columnwidth} p{0.25\columnwidth}}
\toprule
\scriptsize\textbf{Node} & \scriptsize\textbf{Combined Score} & \scriptsize\textbf{Noise Contribution} \\
\midrule
\scriptsize PROFIT\_MARGIN & 0.8526 & 0.0379 \\
\scriptsize NET\_SALES     & 0.8267 & 0.0998 \\
\scriptsize DISCOUNT       & 0.8479 & 0.1241 \\
\bottomrule
\end{tabular}
\vskip 8pt
\noindent{\normalsize \textbf{Table 3:} Node Statistics for 2023-06-10.}
\end{minipage}

\medskip

\paragraph{The pathway extracted for 2023-06-10 comprises four nodes:}
 \texttt{PROFIT\_MARGIN}, \texttt{PROFIT}, \texttt{COST\_OF\_GOODS\_SOLD}, and \texttt{UNIT\_COST}. All nodes exhibit high Combined Scores (ranging from 0.8381 to 0.9086), which confirms that each component strongly contributes to the anomaly signal. Notably, the Noise Contribution progressively increases along the causal chain—from 0.0382 at \texttt{PROFIT\_MARGIN} and 0.0409 at \texttt{PROFIT} to 0.1247 at \texttt{COST\_OF\_GOODS\_SOLD}, reaching a peak of 0.2496 at \texttt{UNIT\_COST}. This trend indicates that the anomaly becomes more pronounced upstream, with \texttt{UNIT\_COST} displaying the most significant deviation. Such a pattern is consistent with the pre-injected \emph{COGSOverstatement} anomaly, thereby confirming that ProRCA has accurately identified the correct causal pathway. \\

\noindent
\begin{minipage}{\columnwidth}
\centering
\label{tab:results_0715}
\renewcommand{\arraystretch}{1.1}
\small
\begin{tabular}{p{0.35\columnwidth} p{0.2\columnwidth} p{0.25\columnwidth}}
\toprule
\scriptsize \textbf{Node} & \scriptsize \textbf{Combined Score} & \scriptsize \textbf{Noise Contribution} \\
\midrule
\scriptsize PROFIT\_MARGIN         & 0.8655 & 0.0382 \\
\scriptsize PROFIT                 & 0.9086 & 0.0409 \\
\scriptsize COGS   & 0.8711 & 0.1247 \\
\scriptsize UNIT\_COST             & 0.8381 & 0.2496 \\
\bottomrule
\end{tabular}
\vskip 8pt
\noindent{\normalsize \textbf{Table 4:} Node Statistics for 2023-07-15.}
\end{minipage}

\medskip

\paragraph{For 2023-09-10, the pathway comprises three nodes}, with the terminal node $\texttt{FULFILLMENT\_COST}$ displaying the highest Combined Score (0.8306) and Noise Contribution (0.2496). This indicates that the anomaly is primarily driven by an abnormal deviation in $\texttt{FULFILLMENT\_COST}$, consistent with the injected FulfillmentSpike anomaly.\\

\noindent
\begin{minipage}{\columnwidth}
\centering
\label{tab:results_0910}
\renewcommand{\arraystretch}{1.1}
\small
\begin{tabular}{p{0.35\columnwidth} p{0.2\columnwidth} p{0.2\columnwidth}}
\toprule
\scriptsize\textbf{Node} & \scriptsize\textbf{Combined Score} & \scriptsize\textbf{Noise Contribution} \\
\midrule
\scriptsize PROFIT\_MARGIN & 0.8166 & 0.0384 \\
\scriptsize PROFIT         & 0.8001 & 0.0326 \\
\scriptsize FULFILLMENT\_COST   & 0.8306 & 0.2496 \\
\bottomrule
\end{tabular}
\vskip 8pt
\noindent{\normalsize \textbf{Table 5:} Node Statistics for 2023-09-10.}
\end{minipage}

\medskip

\paragraph{For 2023-12-10, the pathway comprises three nodes}, with the terminal node \texttt{RETURN\_COST} displaying the highest Combined Score (0.8654) and Noise Contribution (0.0831). This succinct trend indicates that the anomaly is primarily driven by an abnormal deviation in \texttt{RETURN\_COST}, consistent with the injected ReturnSurge anomaly.\\

\noindent
\begin{minipage}{\columnwidth}
\centering
\label{tab:results_1030}
\renewcommand{\arraystretch}{1.1}
\small
\begin{tabular}{p{0.35\columnwidth} p{0.2\columnwidth} p{0.2\columnwidth}}
\toprule
\scriptsize\textbf{Node} & \scriptsize\textbf{Combined Score} & \scriptsize\textbf{Noise Contribution} \\
\midrule
\scriptsize PROFIT\_MARGIN & 0.8098 & 0.0382 \\
\scriptsize PROFIT         & 0.8318 & 0.0390 \\
\scriptsize RETURN\_COST   & 0.8654 & 0.0831 \\
\bottomrule
\end{tabular}
\vskip 8pt
\noindent{\normalsize \textbf{Table 6:} Node Statistics for 2023-10-30.}
\end{minipage}

\medskip

\subsubsection{Discussion of \texttt{min\_score\_threshold}}
When using a min\_score\_threshold of 0.3, the analyzer increased its sensitivity by admitting nodes with lower Combined Scores, thereby identifying a larger set of candidate causal pathways.

For example, on 2023-01-10, five potential pathways were detected, yet the most significant pathway (PCS = 0.0869) correctly linked \texttt{PROFIT\_MARGIN} $\rightarrow$ \texttt{NET\_SALES} $\rightarrow$ \texttt{DISCOUNT}, in agreement with the pre-injected \emph{ExcessiveDiscount} anomaly. Similarly, on 2023-06-10 the pathway with the highest PCS (0.1747) traversed from \texttt{PROFIT\_MARGIN} through \texttt{PROFIT} to \texttt{COST\_OF\_GOODS\_SOLD} and \texttt{UNIT\_COST}, consistent with the expected \emph{COGSOverstatement} anomaly. 

For 2023-09-10, among four candidate pathways, the dominant one (PCS = 0.17) accurately implicated \texttt{FULFILLMENT\_COST}, matching the injected \emph{FulfillentSpike} anomaly. For 2023-12-10, among four candidate pathways, the dominant one (PCS = 0.0578) accurately implicated \texttt{RETURN\_COST}, matching the injected \emph{ReturnSurge} anomaly. 

Although lowering the threshold yielded additional, lower-significance paths, the primary causal pathways remained unchanged, thereby confirming that ProRCA reliably identifies the true root causes despite increased sensitivity.


\section{Conclusion and Key Considerations}
\label{sec:conclusion}

This paper introduced ProRCA, a python package built on top of the DoWhy causal inference library designed to trace multi-hop causal pathways in complex systems. Unlike conventional methods that rely on correlation or provide limited causal insights, ProRCA integrates structural anomaly scoring, noise-based attribution, and depth-first search to systematically reconstruct causal chains from observed effects to their root causes.

Experiments on a synthetic retail dataset with injected anomalies demonstrate ProRCA’s effectiveness in identifying true underlying causes. The package accurately isolated key drivers and exhibited robustness, maintaining stable causal pathways even under reduced sensitivity settings.

While this study validates ProRCA in a controlled environment, several directions remain for future work. Extending the package to real-time streaming applications is crucial for operational deployment. Additionally, ProRCA's accuracy is contingent on the correctness of the DAG, making causal discovery an important area for further research. We propose exploring large language models (LLMs) to enhance automated DAG construction.

Overall, ProRCA advances root cause analysis by offering a comprehensive and explainable approach to multi-hop causal reasoning. Its ability to generate actionable insights in complex, data-intensive environments makes it a valuable tool across industries.

\bibliographystyle{plainnat}  

\begin{thebibliography}{10}

\bibitem{liu2021microhecl}
Li~Liu, Dewei, Chuan He, Xin Peng, Fan Lin, Chenxi Zhang, Shengfang Gong,
  Ziang Li, Jiayu Ou, and Zheshun Wu.
\newblock Microhecl: High-efficient root cause localization in large-scale
  microservice systems.
\newblock In {\em 2021 IEEE/ACM 43rd International Conference on Software
  Engineering: Software Engineering in Practice (ICSE-SEIP)}, pages 338--347.
  IEEE, 2021.

\bibitem{wang2023interdependent}
Dongjie Wang, Zhengzhang Chen, Jingchao Ni, Tong Liang, Zheng Wang, Yanjie
  Fu, and Haifeng Chen.
\newblock Interdependent causal networks for root cause localization.
\newblock In {\em Proceedings of the 29th ACM SIGKDD Conference on Knowledge
  Discovery and Data Mining}, pages 5051--5060, 2023.

\bibitem{wang2023incremental}
Dongjie Wang, Zhengzhang Chen, Yanjie Fu, Yanchi Liu, and Haifeng Chen.
\newblock Incremental causal graph learning for online root cause analysis.
\newblock In {\em Proceedings of the 29th ACM SIGKDD Conference on Knowledge
  Discovery and Data Mining}, pages 2269--2278, 2023.

\bibitem{strobl2022identifying}
Eric~V Strobl and Thomas~A Lasko.
\newblock Identifying patient-specific root causes with the heteroscedastic
  noise model.
\newblock {\em arXiv preprint arXiv:2205.13085}, 2022.
\newblock Accepted at the Journal of Biomedical Informatics.

\bibitem{okati2024root}
Nastaran Okati, Sergio~Hernan Garrido~Mejia, William~Roy Orchard, Patrick
  Bl{\"o}baum, and Dominik Janzing.
\newblock Root cause analysis of outliers with missing structural knowledge.
\newblock {\em arXiv preprint arXiv:2406.05014}, 2024.

\bibitem{budhathoki2021distribution}
Kailash Budhathoki, Dominik Janzing, Patrick Bloebaum, and Hoiyi Ng.
\newblock Why did the distribution change?
\newblock In {\em International Conference on Artificial Intelligence and
  Statistics}, pages 1666--1674. PMLR, 2021.

\bibitem{blobaum2022dowhy}
Patrick Bl{\"o}baum, Peter G{\"o}tz, Kailash Budhathoki, Atalanti~A
  Mastakouri, and Dominik Janzing.
\newblock Dowhy-gcm: An extension of dowhy for causal inference in graphical
  causal models.
\newblock {\em arXiv preprint arXiv:2206.06821}, 2022.

\bibitem{chen2023automatic}
Yinfang Chen, Huaibing Xie, Minghua Ma, Yu~Kang, Xin Gao, Liu Shi, Yunjie Cao,
  Xuedong Gao, Hao Fan, Ming Wen, et~al.
\newblock Automatic root cause analysis via large language models for cloud
  incidents.
\newblock In {\em Proceedings of the 2023 ACM SIGPLAN International Symposium
  on New Ideas, New Paradigms, and Reflections on Programming and Software},
  pages 130--144, 2023.

\bibitem{jeyakumar2019explainit}
Vimalkumar Jeyakumar, Omid Madani, Ali Parandeh, Ashutosh Kulshreshtha,
  Weifei Zeng, and Navindra Yadav.
\newblock Explainit!– a declarative root-cause analysis engine for time
  series data (extended version).
\newblock {\em arXiv preprint arXiv:1903.08132}, 2019.

\bibitem{strobl2024identifying}
Eric~V Strobl and Thomas~A Lasko.
\newblock Identifying patient-specific root causes with the heteroscedastic
  noise model.
\newblock {\em Journal of Biomedical Informatics}, 150:104585, 2024.

\bibitem{han2023root}
Xiao Han, Lu~Zhang, Yongkai Wu, and Shuhan Yuan.
\newblock On root cause localization and anomaly mitigation through causal
  inference.
\newblock In {\em Proceedings of the 32nd ACM International Conference on
  Information and Knowledge Management}, pages 699--708, 2023.
  
\bibitem{wu2020microrca}
Li~Wu, Johan Tordsson, Erik Elmroth, and Odej Kao.
\newblock Microrca: Root cause localization of performance issues in
  microservices.
\newblock In {\em 2020 IEEE/IFIP Network Operations and Management Symposium
  (NOMS)}, pages 1--9. IEEE, 2020.

\bibitem{gong2024porca}
Chang Gong, Di~Yao, Chengyang Luo, Wenbin Li, Jingping Bi, Jin Wang, and
  Yongtao Xie.
\newblock Porca: Root cause analysis with partially observed data.
\newblock {\em ACM Transactions on Knowledge Discovery from Data}, 18(4):1--27,
  2024.

\end{thebibliography}

\end{multicols}
\end{document}